\documentclass{article}
\usepackage{graphicx} 
\usepackage{fullpage}
\usepackage{authblk}
\usepackage{url}
\usepackage{amsmath}
\usepackage{amsfonts}
\usepackage{tabularx}
\usepackage{siunitx}
\usepackage{booktabs}
\usepackage{xcolor}

\title{Giraffe: Adventures in Expanding Context Lengths in LLMs}
\author{Arka Pal\footnote{Correspondence to \texttt{arka@abacus.ai}.}, Deep Karkhanis, Manley Roberts,\\ Samuel Dooley, Arvind Sundararajan, Siddartha Naidu}
\affil{Abacus.AI}
\date{}

\begin{document}

\maketitle

\begin{abstract}
Modern large language models (LLMs) that rely on attention mechanisms are typically trained with fixed context lengths which enforce upper limits on the length of input sequences that they can handle at evaluation time. To use these models on sequences longer than the train-time context length, one might employ techniques from the growing family of \textit{context length extrapolation} methods --- most of which focus on modifying the system of positional encodings used in the attention mechanism to indicate where tokens or activations are located in the input sequence. We conduct a wide survey of existing methods of context length extrapolation on a base LLaMA or LLaMA 2 model, and introduce some of our own design as well --- in particular, a new \textit{truncation} strategy for modifying the basis for the position encoding.

We test these methods using three new evaluation tasks (FreeFormQA, AlteredNumericQA, and LongChat-Lines) as well as perplexity, which we find to be less fine-grained as a measure of long context performance of LLMs. We release the three tasks publicly as datasets on HuggingFace. We discover that linear scaling is the best method for extending context length, and show that further gains can be achieved by using longer scales at evaluation time. We also discover promising extrapolation capabilities in the truncated basis. To support further research in this area, we release three new 13B parameter long-context models which we call {\bf Giraffe}: 4k and 16k context models trained from base LLaMA-13B, and a 32k context model trained from base LLaMA2-13B. We also release the code to replicate our results.\footnote{Github repo at: \protect\url{https://github.com/abacusai/Long-Context}.}

\end{abstract}

\section{Introduction}

In recent years, transformers \cite{vaswani2023attention} have become the dominant neural network architecture in a variety of natural language modelling tasks \cite{hendrycks2021measuring, chen2021evaluating}, by dint of their flexibility and their amenability to being trained on extremely large datasets \cite{gao2020pile, together2023redpajama}. Subsequently, a popular term that has been adopted for these neural networks is `Large Language Models' (LLMs) --- with the `Large' referring both to the training dataset size as well as their parameter count (and indeed, the associated training and environmental cost).

A key element of the standard transformer architecture is its inherent insensitivity to the ordering of the input elements. Attention is naturally a set-like operation in which the position of the elements does not matter \cite{vaswani2023attention}. However, the order of elements is crucial for many important tasks such as parsing natural language, coding, forecasting, etc. Thus it is necessary to inject positional information into the inputs of the LLM, typically in the form of positional encodings.

One possible desideratum of a positional encoding scheme is \textit{context length extrapolation}: the ability to use the LLM for inference on input lengths longer than those it was trained on. Due to the quadratic complexity growth of the attention mechanism in transformers, it is often infeasible to train on large context lengths. The benefit of increased context length is diverse - allowing reading longer documents and papers, more internal consistency in long conversations with users in LLM-powered chatbots, working on bigger codebases, and so on. We can break context length extrapolation down into two main paradigms. First, there is {\it finetuned extrapolation} where a model previously pretrained on shorter contexts is allowed to finetune, or update model weights based on the longer context length. Additionally, there is {\it zero-shot extrapolation} where a model previously pretrained on short contexts is immediately evaluated on longer context lengths with the same weights as the shorter context model.

In this paper, we focus primarily on {\it zero-shot extrapolation} and make the following {\bf key contributions}:

\paragraph{Benchmark of different context extrapolation schemes}  We conduct a survey of methods for context length extrapolation with a pretrained base model, and try a few of our own inventions as well. In particular, we present a new \textit{truncated} basis for position encodings. The focus in this paper on pretrained models is also different from other work in the literature \cite{press2022train, sun2022lengthextrapolatable}, which tend to instead train from scratch with a chosen positional encoding scheme. As mentioned above, although LLMs have been successful, training them is a costly enterprise. Well known closed-source model include GPT-4 \cite{openai2023gpt4} and Claude \cite{claudeanthropic}. Recently the open-source LLaMA \cite{touvron2023llama} has been released by a team at Meta AI, and this was followed by the improved LLaMA2 \cite{touvron2023llama2}. In our view, the resources required to train competitive base models of this nature will remain constrained to a few large players. Therefore, it is imperative to be able to modify the models as desired for the end user---ideally, with a fraction of the compute power applied.

\noindent Our main findings are that:

\begin{itemize}
    \item Linear interpolation is the best as a context length extrapolation method.
    \item All context length extrapolation methods show degradation on task accuracy, even for lengths where they provide otherwise coherent output (and perplexity scores are still reasonable).
    \item Further context length increase can be achieved by utilising a higher scale factor at evaluation time than finetune time, but seemingly only up to a factor of 2x.
\end{itemize}

\paragraph{Public release of LLM weights and evaluation datasets} We release the weights of two new 13B models trained from base LLaMA with an extended context length of 16k \footnote{\url{https://huggingface.co/abacusai/Giraffe-v1-delta-13b-scaled-16}} and a context length of 4k\footnote{\url{https://huggingface.co/abacusai/Giraffe-v1-delta-13b-scaled-4}} on HuggingFace. We also release a 13B model trained to a length of 32k from base LLaMA 2\footnote{\url{https://huggingface.co/abacusai/Giraffe-v2-13b-32k}}. We call this family of models \textbf{Giraffe}. In addition, we release three datasets (LongChat-Lines\footnote{\url{https://huggingface.co/datasets/abacusai/LongChat-Lines}}, FreeFormQA\footnote{\url{https://huggingface.co/datasets/abacusai/WikiQA-Free_Form_QA}} and AlteredNumericQA\footnote{\url{https://huggingface.co/datasets/abacusai/WikiQA-Altered_Numeric_QA}}) to evaluate long context performance of these, and other, models. LongChat-Lines is a key-value fine-grained retrieval task. FreeFormQA and AlteredQA are question-answering datasets based on the Natural Questions Dataset \cite{naturalquestions}. Some existing work \cite{press2022train, sun2022lengthextrapolatable} focuses only on perplexity on a document corpus evaluation set as their measure of extrapolation performance. We find that perplexity scores are not as sensitive a measure of long context performance as our introduced tasks.

\section{Related Work}
\label{sec:related-work}

\paragraph{RoPE} In this work, we examine the efficacy of the positional encoding choice of LLaMA \cite{touvron2023llama} for context lengths longer than the base model was trained on. The positional encoding used by LLaMA is RoPE (Rotary Position Embedding) \cite{su2022roformer}. RoPE works by rotating slices of the query and key projection matrices at different speeds. Thus for example even if the query and key are projected to the same encoding, they will be rotated by different amounts depending on their position in the sequence. If they are subsequently unaligned, their dot product will be smaller relative to what it would be if they were not rotated at all. Conversely, they could become \textit{more} aligned, leading to a larger dot product and attention score. In RoPE, this rotation is happening at different speeds on all 2-slices of the query and key in the embedding dimension, allowing the model to build a complex function of attention scores over distances. One of the main appeals of utilising the RoPE method is that it ensures mathematically that the attention score function is dependent only on the \textit{relative} distance between a query and a key, rather than their absolute positions. This is considered to be a desirable property of LLMs \cite{su2022roformer, shaw2018selfattention}.

\paragraph{ALiBi} Although RoPE was successful in this aim, the work on ALiBi \cite{press2022train} demonstrated that RoPE was not able to perform zero-shot context length extrapolation. The ALiBi paper showed that RoPE quickly degraded as it was tested on context lengths longer than the model had seen during training; it also introduced its own proposed alternative that showed superior extrapolation ability on their benchmarks. However, ALiBi has its own shortcomings; its use of simple linear functions for modulating the attention scores over distance means that it cannot represent as complex distance-attention-functions as the Fourier basis of RoPE. In addition, ALiBi uses a single such function per head, further reducing expressive power. This may explain why, although ALiBi does extrapolate, models which utilize it have worse performance than RoPE-based models on benchmarks such as MMLU \cite{hendrycks2021measuring} and the LMSys arena which measures human preferences \cite{lmsysarena}.

\paragraph{xPos} Sun et al. \cite{sun2022lengthextrapolatable} examine why RoPE fails to extrapolate successfully and determines that this is due to the effect of the high frequency components causing residual noise in the attention score even when tokens are long distances apart. They attempt to address this by adding an exponentially decaying amplitude term to RoPE. This new method, called xPos, decays these noisy high frequency components faster than low frequency components. This method shows good results on from-scratch training of LLMs \cite{sun2022lengthextrapolatable}, and the intuition driving it aligns with our own hypotheses on the deficiency of RoPE. However, Sun et al. do not experiment in our setting of interest: taking a model pretrained with RoPE and seeing if it can be coaxed (via limited finetuning) to learn the xPos encoding instead. Furthermore, their experiments demonstrate that Blockwise Causal Attention is necessary for them to achieve extrapolation.

\paragraph{Linear Scaling/Positional Interpolation}

This simple but effective context length extrapolation technique was concurrently reported by kaiokendev \cite{kaiokendev} and by a team at Meta \cite{chen2023extending}. The method that is used here is to simply divide the position vector by a scaling factor which fits the input within the context length of the original model. The intuition of this technique is to utilize the LLM's interpolation capability, rather than relying on extrapolation. It is a well known phenomenon that neural networks tend to interpolate within a range of previously seen values better than they extrapolate outside that range (e.g. \cite{xu2021neural}). In the specific case of positional encodings,  \cite{chen2023extending} claim that positional interpolation avoids the risk of massive numerical explosion in attention values associated with extrapolation. We perform many experiments on this scheme and variations of it and report the results in this paper.

\paragraph{Randomized Positional Encodings}
Ruoss et al present this method in \cite{ruoss2023randomised}. During training they randomly generate their position vector by drawing N many samples uniformly without replacement from the range [1, L], where N is the training context length and L is a large value that is greater than the (assumed to be known prior) maximum evaluation context length. These sampled positions are then sorted in increasing order and act as the position inputs that the model sees at evaluation time. During evaluation, the position inputs [1, ..., M] are given to the model. The authors claim improved performance on context length extrapolation. We independently arrived at a similar idea to this paper but instead randomized by drawing from sub-integer positions approximately in the range [1, N]; see Section \ref{sec:experiment-setup} for further details. We also note that Ruoss et al investigate the use of such a scheme for training LLMs from scratch, whereas we are primarily interested in post-hoc finetuning a pretrained LLM with randomization to enable context length extrapolation.

\section{Assessing Long Context Extrapolation}
\label{sec:tasks}

The main question posed in this paper revolves around extending the context length capacity of LLMs. To evaluate this, a commononly used metric in the literature is {\it perplexity}  \cite{press2022train, sun2022lengthextrapolatable, su2022roformer}. However, as we show in Section~\ref{sec:perplexity}, perplexity is somewhat coarse-grained for evaluating how well the model can use longer context windows. Our intuition is that --- in many natural language datasets --- a reasonable perplexity score can be achieved even if the model is only attending to information in a limited range (the final 512 tokens, say) of the context window. For example, a positional encoding scheme which simply masks out any elements of the context (and inner key and query activations in the attention heads) that are greater in length than what the model was trained on should succeed in achieving a reasonable perplexity score, but would fare poorly on the tasks we describe below.

We expand upon existing work to look at the {\it accuracy} of a model when presented with problems which have verifiable answers. In using this metric, we can evaluate how the model is using the additional contextual information in order to respond to prompts. We rely on two types of evaluation tasks to assess models' ability to extract and use information from long input contexts: the first is key-value retrieval tasks and the other is question answering tasks.  By using these two types of tasks, we enforce the requirement of the model to attend to the full context in order to obtain high accuracies. We consider the retrieval task to be a more pure test of information retrieval free of many natural language biases. However, the retrieval task is a somewhat artificial construct which the LLM will likely not have seen during training, so we also include question answering to replicate more real-world tasks.

\paragraph{LongChat-Lines} We start with a synthetic fine-grained key-value retrieval task first proposed in \cite{littleretrieval} and also used by \cite{longchat2023}. While these works are excellent given the standard contexts of LLMs, they lack the longer context lengths that are needed to evaluate our experiments. Thus, we utilize the same task as \cite{littleretrieval,longchat2023}, but generate additional samples of longer context lengths. This task gives the model a prompt with lines of the form:
\begin{itemize}
    \item line grotesque-classmate: REGISTER\_CONTENT is $<42527>$
    \item line imperfect-bull: REGISTER\_CONTENT is $<3119>$
    \item line supreme-inversion: REGISTER\_CONTENT is $<13960>$
    \item ...
\end{itemize}
The model is asked to memorize the value corresponding to the REGISTER\_CONTENT for each line and is asked at the end to retrieve the value for a specific line. By varying the number of lines in the prompt, we can control the context length. We release longer length versions of this task than in \cite{littleretrieval}, and we also release the generation script for this task.

\paragraph{WikiQA} We also create two new datasets from the Natural Questions \cite{naturalquestions} dataset with longer context evaluations specifically in mind which we collectively term WikiQA. In this evaluation, the prompt given to the LLM is in the format of a Wikipedia document followed by a question pertaining to that document; the model is asked to answer the question. We ensure the answer to the question is a short answer which is either a single word or a small sentence that has an exact string match in the document given to the LLM as input. We call this task {\bf Free Form QA (FFQA)}.

A potential issue in a Wikipedia based dataset however is that the model could perhaps correctly answer from its pretrained corpus and not specifically using the information in the context. To resolve this, we have created another “altered” dataset, which we call {\bf Altered Numeric QA (AltQA)}. This dataset consists only of questions which have numerical answers. Here, we change the answer and every occurrence of the answer in the document to a different number, thus ensuring that the LLM must attend to the context, and only the context, in order to give a correct answer. The modification is made as follows:

\begin{itemize}
    \item If the answer is a year, which is quite frequent, (i.e. it is between 1000-2100), we change it to a different random value within +/- 10 of the original value. We treat years as a special case so as not to disrupt the overall coherence of the document by having highly anachronistic date values.
    \item If the answer is any other number, we change it to a different random number which has the same number of digits.
\end{itemize}

\begin{figure}
\includegraphics[width=\textwidth]{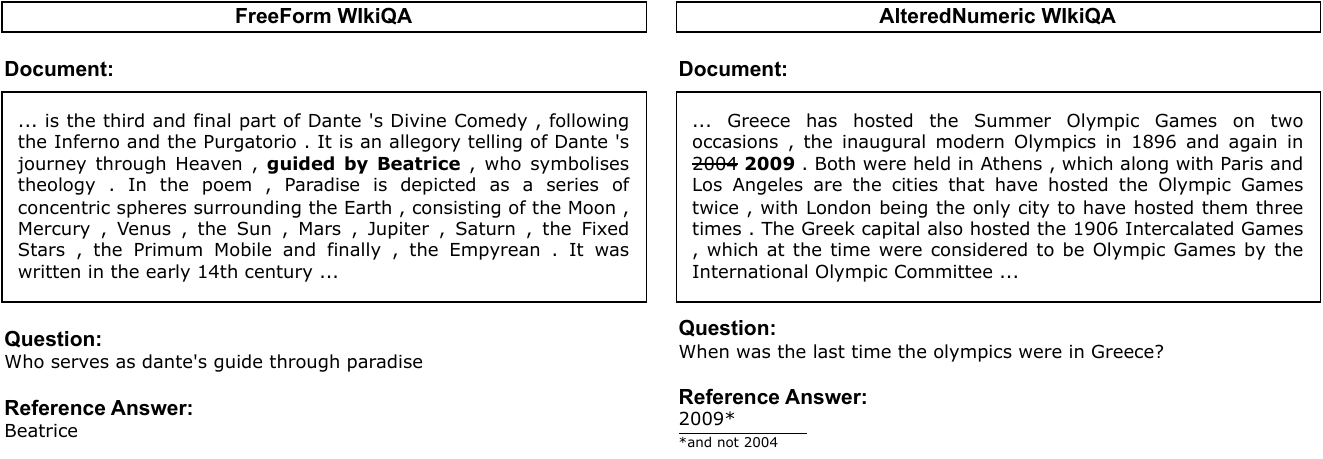}    
\caption{Example QA snippets from our WikiQA dataset.}
\centering
\label{fig:wikiqa_eg}
\end{figure}

Figure \ref{fig:wikiqa_eg} highlights examples from our WikiQA dataset. Since the contexts in our application are long, the location of the answer within the context could play a significant role in the model's ability to answer the question. Therefore, we utilize both of the WikiQA tasks to conduct analysis on the performance of the LLM as both the answer location moves within the document (in the beginning 10\%, the last 10\%, or randomly anywhere else), as well as with the question given at the beginning or the end of the prompt --- in a bid to replicate the analyses of \cite{liu2023lost}.

\section{Context Length Extrapolation Techniques}
\label{sec:experiment-setup}

We examine several context length extrapolation techniques, including existing approaches (or slight variations on them) as well as our own newly proposed approaches.

\subsection{Existing Context Length Extrapolation Techniques}

Several methods exist to adapt RoPE positional encodings to longer context lengths. We evaluated the following techniques.

\paragraph{Linear Scaling/Positional Interpolation}

Here, the position vector is divided by a scaling factor. Hence if the original model was trained on a range of positions $[0, 1, ..., 2048]$, say, then the new model will see instead $[\frac{0}{x}, \frac{1}{x}, ..., \frac{2048}{x}]$ where $x$ is the scaling factor.

\paragraph{xPos}

We wanted to examine whether a checkpoint trained with the base model's RoPE encoding scheme could be finetuned to the xPos \cite{sun2022lengthextrapolatable} scheme.
On top of the programming hurdle of patching the entire attention module to handle xPos' unique transformation of keys and queries, the major issue presented by this sort of adaptation is xPos' sensitivity to floating point precision. The method relies on scaling the key by numeric values with large (absolute) exponents; these later cancel in the dot product with the query. For long contexts, however, the large values can actually exceed the magnitude supported by float16. We chose to work around this by performing the core attention operation in float32 at the cost of a 2X training slow down.

\paragraph{Randomized Position Encodings}

Here we randomize the distances between the position values uniformly in the range $[\epsilon, 2]$ for $0<\epsilon\ll1$, rather than using the typical $[0, 1, ..., n]$ which has fixed intervals of size 1. The intuition behind this approach is that by showing the model many different intra-position distances at finetuning time, the model will be able to generalize to any choice of fine-grained positions at evaluation time, thereby allowing for an effective increase in context length by choosing smaller divisions. This has some similarity to the procedure described in Ruoss et al. \cite{ruoss2023randomised}. We set an upper bound of 2 so that the model will in expectation see a final position of n (as $\mathbb{E}[X] \approx 1$ for $X\sim U(\epsilon, 2)$). We also set a positive, non-zero lower bound of $\epsilon$ in order to avoid issues with position aliasing due to limited numerical precision.

\subsection{Newly Proposed Context Length Extrapolation Techniques}
    
\paragraph{Power Scaling}

\begin{figure}[!t]
    \centering
    \includegraphics[width=0.65\textwidth]{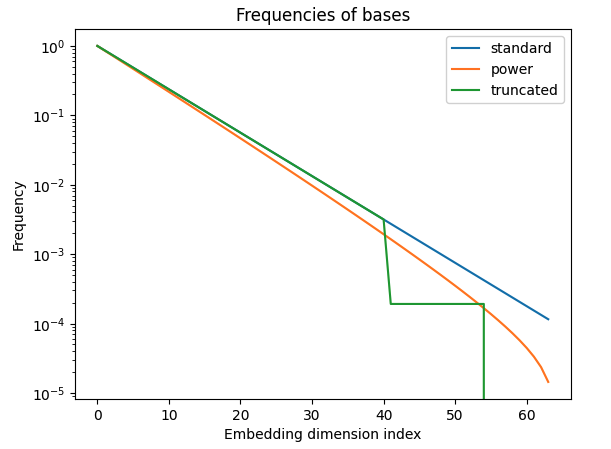}
    \caption{Comparison of the standard RoPE basis vs the power basis and the truncated basis. The x-axis spans over the embedding dimension, and the y-axis is the frequency value of the sine-cosine basis.}
    \label{fig:bases}
\end{figure}

In the original RoPE, the basis that is used is given by:
    \begin{equation}\label{eq:ropebasis}
    \Theta = \{\theta_i = 10000^{-\frac{2(i-1)}{d}}\mid i \in \{1, 2, \ldots, \frac{d}{2}\}\}
    \end{equation}\label{eq:powerbasis}
    where $d$ is the embedding dimension. We use instead the basis given by:
    \begin{equation}
    \Theta^* = \left\{\theta_i^* = \theta_i\left(1-\frac{2i}{d}\right)^{k}\mid i \in \{1, 2, \ldots, \frac{d}{2}\}\right\}
    \end{equation}
    where $k$ is a parameter to be set. By applying this transformation, the high frequency (short distance) elements of the basis are less affected than the low frequency (long distance) elements, which are made even lower in frequency -- see Figure \ref{fig:bases}. By doing so, our hope was that the model would have to perform less complex extrapolation for the low frequencies where it has not seen the full range of the periodic function during train time, and thereby extrapolate better. A potential issue however is that the model relies on specific relationships across frequencies that a linear transform preserves but a non-linear transformation destroys.

\paragraph{Truncated Basis}

Beginning from Equation \ref{eq:ropebasis}, we instead use the basis given by applying:
    \begin{equation}\label{eq:truncatedbasis}
    \theta^*_i = 
    \begin{cases} 
    \theta_i & \text{for } \theta_i \ge b, \\
    \rho & \text{for } a < \theta_i < b, \\
    0 & \text{for } \theta^*_i \le a.
    \end{cases}
    \end{equation}
    Where $\rho$ is a fixed value that is relatively small, and $a$ and $b$ are chosen cutoff values. The idea here is that we wish to preserve the high frequency components of the basis but set the low frequency elements to a constant value---in this case, 0. By doing so with a judicious choice of cutoff $a$, the model will have experienced all values of the basis in the context length used during finetuning (due to the periodic nature of the sine and cosine functions) and should therefore extrapolate better to larger context lengths for evaluation.
    However, the model still needs to be able to distinguish between distances that span the entire context it was trained on, so we include the $\rho$ fixed frequency as well. In summary, we hope that with this basis the model can avoid the issue of having to learn complicated coefficients in the entire RoPE basis by instead learning smooth functions at longer distances (as demonstrated in the paper \cite{chen2023extending}).

    In Figure \ref{fig:bases}, we visually compare the frequencies produced by the standard RoPE basis, power scaling, and truncation.

\section{Results \& Discussion}
\label{sec:results}

In the following experiments, we finetuned a base LLaMA-13B model on a portion of the RedPajama dataset \cite{together2023redpajama} which has been modified so that each data sample has a size of exactly 4096 tokens. We trained with each positional encoding approach until the evaluation loss roughly plateaued. Loss curves can be found in Appendix \ref{sec:loss-curves}.

We then further applied instruction finetuning (IFT) with the Vicuna dataset \cite{zheng2023judging} and using LoRA \cite{hu2021lora} on the base model. However, we discovered that although IFT did boost accuracies on LongChat-Lines, it did not significantly change the range of contexts that the base model was able to deal with (see Figure \ref{fig:impact_of_ift} in Appendix \ref{sec:impact-of-ift}). This we found to be a marked contrast with the WikiQA variants; there, IFT was necessary for the model to produce any meaningful results at all. Hence for LongChat-Lines, we used non-IFT models; for WikiQA, we performed evaluation on a subset of the more promising models with additional IFT.

\subsection{Finetuned Context Length Extrapolation}

\newcolumntype{Y}{>{\centering\arraybackslash}X}

\begin{table}[!t]
\small
\centering
\begin{tabularx}{\textwidth}{Y|Y|Y|Y|Y|Y|Y}

Context Length & Linear Scaling (Factor=4) & Linear Scaling (Factor=16) & Power basis & Truncated basis & Randomized position & xPos \\
\hline
\hline
2500 & 0.7 & 0.64 & 0.96 & 0.42 & 0.54 & 0 \\
3600 & 0.64 & 0.42 & 0.64 & 0.26 & 0.54 & 0 \\
4200 & 0.56 & 0.56 & 0.1 & 0.18 & 0.3 & 0 \\
4800 & 0.66 & 0.62 & 0 & 0.14 & 0.14 & 0 \\
7100 & 0.36 & 0.4 & 0 & 0.04 & 0.16 & 0 \\
9400 & 0 & 0.22 & 0 & 0 & 0 & 0 \\
11800 & 0 & 0.14 & 0 & 0 & 0 & 0 \\
14000 & 0 & 0.12 & 0 & 0 & 0 & 0 \\
16000 & 0 & 0.1 & 0 & 0 & 0 & 0 \\
17500 & 0 & 0 & 0 & 0 & 0 & 0 \\
20000 & 0 & 0 & 0 & 0 & 0 & 0 \\
22000 & 0 & 0 & 0 & 0 & 0 & 0 \\

\end{tabularx}
\caption{After finetuning LLaMA-13B with a base context length of 4096, this table represents evaluations with different context extrapolation methods on LongChat-Lines. An accuracy of 1.0 would indicate perfect performance and 0.0 indicates getting every evaluation wrong. The power basis uses a parameter of $k= 0.5$. The truncated basis uses the following parameters: $a = \frac{1}{8}\frac{2\pi}{2048}, b = \frac{2\pi}{2048}, \rho=\frac{1}{16}\frac{2\pi}{2048}$. Randomization uses a lower bound parameter of $\epsilon = \frac{1}{16}$. Evaluations are all performed without additional instruction finetuning.}
\label{table:finetuned-fine-grained-retrieval}
\end{table}

\paragraph{LongChat-Lines}

We conducted evaluations on LongChat-Lines with the techniques described in Section \ref{sec:experiment-setup} and report the results in Table \ref{table:finetuned-fine-grained-retrieval}. We expected all models to be able to perform with a non-zero accuracy until at least 4200 given that the model is finetuned on context lengths of 4096 and convergence is achieved in all cases. However, this turned out not to be the case for xPos, which was not able to perform the task at all. We suspect this may be because xPos is too different from the RoPE basis for the model to be able to adapt in finetuning; as we see in Appendix \ref{sec:loss-curves}, the training and evaluation loss for xPos was not able to reach the same values as the other methods. This may also be a product of the numerical precision issues that are encountered in the implementation of xPos.

Linear scaling is able to achieve successful context length extrapolation. It is worth mentioning here that we would expect scaling with a factor of $x$ to achieve non-zero accuracies up to $2048\cdot x$, due to the base model being trained on a context length of 2048. Although this is observed with linear scaling with a factor of 4, we see in Table \ref{table:finetuned-fine-grained-retrieval} much quicker degradation as context length increases with a scaling factor of 16. By context length \num{17500} it is already recording 0\% accuracy even though we naively would expect reasonable performance up to roughly \num{32000} context length. We believe that this indicates that there are limits to the interpolation methodology and are interested in examining this further in future work.

The power basis, although it performs best at the shortest context, also decays fastest and is unable to show any extrapolation performance beyond 4200 context at all.

The randomized position approach may appear to be extrapolating based on the results in the table. However, this is likely due to how we evaluated the model. At train time, the model samples distances uniformly in $[\epsilon, 2]$ as described in Section \ref{sec:experiment-setup}. At evaluation time, it is unclear a priori what the best choice of positions is. We tried a range of different approaches: fixed distances of size 1, uniform random in  $[\epsilon, 2]$ and uniform random in $[\epsilon, 1]$. We found best results for extrapolation with the latter, so we report this. We hoped that by reducing the upper bound further, we could coax the desired context length extrapolation from the model. However, going to $[\epsilon, 0.5]$ and below significantly degraded the performance of the model. Our conclusion from this is that the model cannot independently learn to represent each position without knowing the other positions as well. An interesting avenue for future work would be to condition the Q-proj and K-proj matrices on the sampled positions during training (and evaluation). 

The truncated basis does seem to offer true context length extrapolation, as it is able to achieve non-zero accuracies on context lengths outside any values it has seen before. Although the performance does degrade as the length increases and the current manifestation of this is inferior in performance to linear scaling, we believe that this may be a direction of investigation that can lead to better extrapolation performance. Truncation can also be combined with linear scaling, as we discuss in Section 5.2.

\paragraph{WikiQA Variants}

We further conducted evaluations of the linear scaling and truncated basis approaches on the WikiQA variants described in Section \ref{sec:experiment-setup}. Unlike the retrieval task, we found that models were unable to perform this task successfully without any instruction finetuning, so we performed this analysis on only a few approaches of interest. The results are shown in Tables \ref{table:finetuned-altered-numeric-qa} and \ref{table:finetuned-freeform-qa}. They largely match the pattern seen in LongChat-Lines---linear scaling with scale factor 4 is able to perform the task up to 7500 context but not beyond, whilst scale factor 16 is able to surpass this cutoff but with a slope-off in accuracy. As with LongChat-Lines, we see that the models appear to show some degradation of accuracy as context length increases. We see again that the truncated basis is able to extrapolate successfully to about the same context length as in LongChat-Lines with comparable accuracies to linear scaling with scale 4, but again seemingly cannot go further than a context length of about 8k.

\begin{table}[!t]
\small
\centering
\begin{tabularx}{\textwidth}{Y|Y|Y|Y}

Context Length & Linear scaling (Factor=4) & Linear scaling (Factor=16) & Truncated basis \\
\hline
\hline
2000 & 0.72 & 0.69 & 0.74 \\
3800 & 0.72 & 0.73 & 0.73 \\
7500 & 0.62 & 0.70 & 0.46 \\
15000 & 0.00 & 0.68 & 0.00 \\
24000 & 0.00 & 0.56 & 0.00 \\
32000 & 0.00 & 0.18 & 0.00 \\

\end{tabularx}
\caption{Accuracy on the AlteredNumeric WikiQA variant of models finetuned with different context extrapolation methods. Evaluations are all performed with instruction finetuning.}
\label{table:finetuned-altered-numeric-qa}
\end{table}

\begin{table}[!t]
\small
\centering
\begin{tabularx}{\textwidth}{Y|Y|Y|Y}

Context Length & Linear scaling (Factor=4) & Linear scaling (Factor=16) & Truncated basis \\
\hline
\hline
1900 & 0.44 & 0.47 & 0.46 \\
3800 & 0.49 & 0.44 & 0.55 \\
7600 & 0.46 & 0.45 & 0.36 \\
15000 & 0.00 & 0.48 & 0.00 \\
24000 & 0.00 & 0.42 & 0.00 \\
32000 & 0.00 & 0.20 & 0.00 \\

\end{tabularx}
\caption{Accuracy on the FreeForm WikiQA variant of models finetuned with different context extrapolation methods. Evaluations are all performed with instruction finetuning.}
\label{table:finetuned-freeform-qa}
\end{table}

\subsection{Zero-Shot Linear Scaling}

In the previous section, we examined the performance of linear scaling using the same value at finetuning time as at evaluation time. In this section, we instead investigate the effect of using a different scaling factor at evaluation time than what the model was trained on. In Table \ref{table:zero-shot-linear-scaling}, we show results for this strategy as evaluated on LongChat-Lines. We find in general that if the model is trained with a scale factor of $x$, then the model can successfully evaluate zero-shot with a scale factor of 2$x$ (with some reduction of accuracy within the range of context lengths the model could previously handle). It also appears that at a scale factor of 16, the model is no longer able to increase its effective context length by using this approach. We also find that evaluating with $>$2$x$ leads to the model breaking and being unable to perform the task.

We show that zero-shot linear scaling can be applied successfully after finetuning with the truncated basis. Interestingly, whilst for linear scaling using a longer scale factor at evaluation time results in a deterioration of the accuracy on context lengths the model could previously handle, this does not appear to be the case for the truncated basis---instead, the range of context lengths that the model achieves non-zero accuracy on increases, and accuracy \textit{improves} among context lengths the model was finetuned on.

\begin{table}[!t]
\centering
\resizebox{\textwidth}{!}{
\begin{tabular}{lcccccccccccc}
\toprule
& \multicolumn{3}{c}{Train Scaling = 1 (base model)} & \multicolumn{3}{c}{Train Scaling = 4} & \multicolumn{3}{c}{Train Scaling = 16} & \multicolumn{3}{c}{Truncated (Scaling = 1)} \\
\cmidrule(lr){2-4} \cmidrule(lr){5-7} \cmidrule(lr){8-10} \cmidrule(lr){11-13}
Context Length & Eval 1 & Eval 2 & Eval 4 & Eval 4 & Eval 8 & Eval 16 & Eval 16 & Eval 32 & Eval 64 & Eval 1 & Eval 2 & Eval 4 \\
\midrule
2500 & 0 & 0.32 & 0 & 0.88 & 0.64 & 0 & 0.64 & 0.24 & 0 & 0.42 & 0.58 & 0 \\
3600 & 0 & 0.3 & 0 & 0.8 & 0.58 & 0 & 0.42 & 0.26 & 0 & 0.26 & 0.44 & 0 \\
4200 & 0 & 0.18 & 0 & 0.86 & 0.62 & 0 & 0.56 & 0.12 & 0 & 0.18 & 0.28 & 0 \\
4800 & 0 & 0 & 0 & 0.86 & 0.62 & 0 & 0.62 & 0.22 & 0 & 0.14 & 0.26 & 0 \\
7100 & 0 & 0 & 0 & 0.64 & 0.38 & 0 & 0.4 & 0.12 & 0 & 0.04 & 0.04 & 0 \\
9400 & 0 & 0 & 0 & 0 & 0.32 & 0 & 0.22 & 0.12 & 0 & 0 & 0.04 & 0 \\
11800 & 0 & 0 & 0 & 0 & 0.30 & 0 & 0.14 & 0.1 & 0 & 0 & 0.04 & 0 \\
14000 & 0 & 0 & 0 & 0 & 0.1 & 0 & 0.12 & 0.04 & 0 & 0 & 0 & 0 \\
16000 & 0 & 0 & 0 & 0 & 0.12 & 0 & 0.1 & 0.02 & 0 & 0 & 0 & 0 \\
17500 & 0 & 0 & 0 & 0 & 0 & 0 & 0 & 0 & 0 & 0 & 0 & 0 \\
20000 & 0 & 0 & 0 & 0 & 0 & 0 & 0 & 0 & 0 & 0 & 0 & 0 \\
22000 & 0 & 0 & 0 & 0 & 0 & 0 & 0 & 0 & 0 & 0 & 0 & 0 \\
\bottomrule
\end{tabular}
}
\caption{
Accuracy on LongChat-Lines of models finetuned with different context extrapolation methods and then evaluated with additional linear scaling. Whenever the evaluation linear scale is greater than the training linear scale, this produces zero-shot context length extrapolation. 
In general, increasing the evaluation context length 2x over train actually does double the usable context length, at an accuracy cost, for train lengths 1 and 4 (for train length 16, it does not). The truncated basis accuracy \textit{improves} with 2x scaling. A more aggressive scale up of 4x times the train length leads to apparent model failure. }
\label{table:zero-shot-linear-scaling}
\end{table}

\subsection{Comparing Perplexity To Tasks} \label{sec:perplexity}

In Section \ref{sec:tasks}, we introduced two tasks which specifically require a long-context LLM to extract answers from throughout the entire text, arguing that these tasks may assess long context performance better than raw perplexity. To analyze how perplexity fares as compared to these tasks, we report perplexities on a held out set of the RedPajama dataset for a subset of our trained models (see Table \ref{table:perplexity}). Perplexity scores do show a large increase when a context length is reached that the model is completely unable to deal with (for example, beyond 2k on the base LLaMA model, or beyond 8k on the linear scale 4 model). However, they are appear less effective for showing the decrease in long-context capability within that effective range. In particular, while we observe a steep slope-off in performance on LongChat-Lines and the WikiQA variants as the context length increases for the linear scale 4 and truncated basis columns of Table \ref{table:finetuned-fine-grained-retrieval}), this degradation is not strongly reflected in the perplexity scores at those contexts. However, the linear scale 16 model does appear to have well-correlated perplexity and accuracy on our tasks. Perhaps most tellingly, we see the shortcoming of perplexity for between-model comparisons. According to Table \ref{table:perplexity}, the truncated basis performs best at 8k and below; however, in Tables \ref{table:finetuned-fine-grained-retrieval}, \ref{table:finetuned-altered-numeric-qa} and \ref{table:finetuned-freeform-qa} truncated is significantly lower in performance compared to the linear scaled models at 8k context.

Perplexity is commonly used in the literature to measure long context performance \cite{su2022roformer, sun2022lengthextrapolatable}, but we believe these results show it is not in itself a sufficient measure of long context performance, but is best utilised with other tasks which additionally probe the capabilities of the LLM.

\begin{table}[h]
\small
\centering
\begin{tabular}{@{}lcccc@{}}
\toprule
Context Length & LLaMA Base & Linear Scaling (Factor=4) & Linear Scaling (Factor=16) & Truncated Basis \\
\midrule
512  & 4.06  & 4.06  & 4.05 & 3.79 \\
1k   & 3.88  & 3.87  & 3.86 & 3.63 \\
2k   & 3.79  & 3.75  & 3.74 & 3.52 \\
4k   & 9022  & 3.66  & 3.66 & 3.46 \\
8k   & 7198  & 3.79  & 3.97 & 3.78 \\
16k  & 5141  & 14902 & 5.43 & 15793 \\
24k  & 4980  & 21236 & 8.73 & 13929 \\
32k  & 4408  & 55480 & 98.12 & 12534 \\
\bottomrule
\end{tabular}
\caption{Perplexity scores on a held out evaluation set of the RedPajama dataset at various context lengths on different models. The evaluation length is 256 tokens, and the prompt given to the models is the previous N - 256 tokens of the document, where N is the context length we evaluated. When the context length is too long for the model to handle effectively, the perplexity does blow up; however, within ranges the model can handle, perplexity appears to be less sensitive to context-usage degradation than the LongChat-Lines task, and does not follow the same between-model ranking as that of LongChat-Lines or WikiQA.
}
\label{table:perplexity}
\end{table}

\subsection{Analysis of question and answer positioning}

For the WikiQA variants, we performed a stratified analysis of the effect of the position of the answer and the question. As described in Section \ref{sec:experiment-setup}, we looked at the impact of placing the answer within the first 10\% of the document, the last 10\%, or elsewhere randomly. We also examined the effect of putting the question at the beginning or the end of the prompt. The results are shown in Tables \ref{table:stratified-altered-numeric-qa} and \ref{table:stratified-freeform-qa}, performed on the model with linear scaling with a factor of 16, with additional instruction finetuning.

We aimed to build on a similar analysis from \cite{liu2023lost}. However, we were not able to replicate the results shown in that paper on the LongChat-13B (16K) model (to which our modeling approach is most comparable). On both FreeFormQA and AlteredNumericQA, we observed no clear trend with regards to the location of the answer within the prompt and the model's accuracy up to 15k context length. There also did not appear to be a significant impact on the location of the question for AlteredNumericQA, but there is a noticeable impact observed for FreeFormQA where having the question at the end appears to have a significant improvement in accuracy. However, at 24k and 32k context lengths we see a clear indication in both datasets for both the answer at the end and question at the end returning superior accuracy to their placements elsewhere. These results are a marked contrast to those in \cite{liu2023lost}. Our take away from this is that there is plausibly a great deal of task-conditional variability in the performance of LLMs with regards to how well they can utilize all portions of the context; even small differences in task construction can lead to large differences in observed trends.

\begin{table}[!t]
\small
\centering
\begin{tabularx}{\textwidth}{Y|Y|Y|Y|Y|Y}
Context Length & \multicolumn{3}{c}{Answer Location} & \multicolumn{2}{c}{Question Location} \\
 & Start & Middle & End & Start & End \\
\hline
\hline
2000 & 0.74 & 0.68 & 0.66 & 0.70 & 0.69 \\
3800 & 0.70 & 0.67 & 0.66 & 0.63 & 0.73 \\
7500 & 0.69 & 0.63 & 0.65 & 0.61 & 0.70 \\
15000 & 0.68 & 0.68 & 0.69 & 0.68 & 0.68 \\
24000 & 0.30 & 0.26 & 0.50 & 0.14 & 0.56 \\
32000 & 0.13 & 0.13 & 0.23 & 0.15 & 0.18 \\
\hline
\end{tabularx}
\caption{Stratified accuracy analysis on the AlteredNumericQA task. For answers, ``Start" refers to the first 10\% of the document, ``End" to the last 10\%, and ``Middle'' to any other location. For questions, ``Start'' refers to placing the question before the rest of the language prompt, and ``End'' refers to placing the question at the end. Results are reported on LLaMA-13B finetuned with a linear scale of 16, with IFT applied.}
\label{table:stratified-altered-numeric-qa}
\end{table}

\begin{table}[!t]
\small
\centering
\begin{tabularx}{\textwidth}{Y|Y|Y|Y|Y|Y}
Context Length & \multicolumn{3}{c}{Answer Location} & \multicolumn{2}{c}{Question Location} \\
 & Start & Middle & End & Start & End \\
\hline
\hline
1900 & 0.37 & 0.40 & 0.36 & 0.27 & 0.47 \\
3800 & 0.40 & 0.30 & 0.32 & 0.24 & 0.44 \\
7600 & 0.35 & 0.34 & 0.35 & 0.24 & 0.45 \\
15000 & 0.44 & 0.43 & 0.45 & 0.40 & 0.48 \\
24000 & 0.18 & 0.20 & 0.40 & 0.10 & 0.42 \\
32000 & 0.07 & 0.11 & 0.18 & 0.04 & 0.20 \\
\hline
\end{tabularx}
\caption{Stratified accuracy analysis on the FreeFormQA task. For answers, ``Start" refers to the first 10\% of the document, ``End" to the last 10\%, and ``Middle'' to any other location. For questions, ``Start'' refers to placing the question before the rest of the language prompt, and ``End'' refers to placing the question at the end. Results are reported on LLaMA-13B finetuned with a linear scale of 16, with IFT applied.}
\label{table:stratified-freeform-qa}
\end{table}

\section{Conclusion and Limitations}

In this paper we examined multiple approaches to finetuning a pretrained base LLaMA and LLaMA2 LLM that has a limited context length such that it is capable of extrapolating zero-shot to new, longer context lengths. We compared the methods using perplexity, as well as two custom tasks that probe long context performance; we find that the custom tasks offer a more fine-grained understanding of long context performance than perplexity. We showed that the method of linear interpolation performed best at context length extrapolation, and found some promise in the potential of using a new basis which we termed the \textit{truncated} basis. We release three models which we call Giraffe that extend the context length of the base LLaMA and LLaMA 2 models using the method of linear interpolation. 

There is significant room for building on the work presented in this paper. We note that all methods show a degradation in accuracy on our evaluation tasks as context length increases, even though perplexity often remains reasonable and the model can still produce coherent outputs. This is a shortcoming that would be of interest to address, and in our view is necessary for claiming `true' long context extrapolation ability of a model.

The limitations of this work are that we only conducted our perplexity analysis on a single document dataset. Future work could look to replicate this analysis on other datasets. Additionally, we focused specifically on context-length extrapolation from a pretrained base model, and in particular the LLaMA and LLaMA 2 models trained with RoPE positional encodings. Future work could investigate whether the analysis herein extends to other positional encoding types and models. Future work could also address the limitations of the linear interpolation method. We see some evidence on the LongChat-Lines task in particular of accuracy degradation as the scale factor is increased. What is the limit of the size of scale factor of this method? Is there a point beyond which it simply does not improve the range of contexts it can handle? Furthermore, can the truncated basis approach which seems to show signs of true extrapolation capability be modified in a manner to gain parity with or surpass the linear interpolation method? We believe these are some potential future directions of interest.

\bibliography{references.bib} 

\begin{thebibliography}{10}

\bibitem{vaswani2023attention}
Ashish Vaswani, Noam Shazeer, Niki Parmar, Jakob Uszkoreit, Llion Jones,
  Aidan~N. Gomez, Lukasz Kaiser, and Illia Polosukhin.
\newblock Attention is all you need, 2023.

\bibitem{hendrycks2021measuring}
Dan Hendrycks, Collin Burns, Steven Basart, Andy Zou, Mantas Mazeika, Dawn
  Song, and Jacob Steinhardt.
\newblock Measuring massive multitask language understanding, 2021.

\bibitem{chen2021evaluating}
Mark Chen, Jerry Tworek, Heewoo Jun, Qiming Yuan, Henrique~Ponde
  de~Oliveira~Pinto, Jared Kaplan, Harri Edwards, Yuri Burda, Nicholas Joseph,
  Greg Brockman, Alex Ray, Raul Puri, Gretchen Krueger, Michael Petrov, Heidy
  Khlaaf, Girish Sastry, Pamela Mishkin, Brooke Chan, Scott Gray, Nick Ryder,
  Mikhail Pavlov, Alethea Power, Lukasz Kaiser, Mohammad Bavarian, Clemens
  Winter, Philippe Tillet, Felipe~Petroski Such, Dave Cummings, Matthias
  Plappert, Fotios Chantzis, Elizabeth Barnes, Ariel Herbert-Voss,
  William~Hebgen Guss, Alex Nichol, Alex Paino, Nikolas Tezak, Jie Tang, Igor
  Babuschkin, Suchir Balaji, Shantanu Jain, William Saunders, Christopher
  Hesse, Andrew~N. Carr, Jan Leike, Josh Achiam, Vedant Misra, Evan Morikawa,
  Alec Radford, Matthew Knight, Miles Brundage, Mira Murati, Katie Mayer, Peter
  Welinder, Bob McGrew, Dario Amodei, Sam McCandlish, Ilya Sutskever, and
  Wojciech Zaremba.
\newblock Evaluating large language models trained on code, 2021.

\bibitem{gao2020pile}
Leo Gao, Stella Biderman, Sid Black, Laurence Golding, Travis Hoppe, Charles
  Foster, Jason Phang, Horace He, Anish Thite, Noa Nabeshima, Shawn Presser,
  and Connor Leahy.
\newblock The pile: An 800gb dataset of diverse text for language modeling,
  2020.

\bibitem{together2023redpajama}
Together Computer.
\newblock Redpajama: An open source recipe to reproduce llama training dataset,
  April 2023.

\bibitem{press2022train}
Ofir Press, Noah~A. Smith, and Mike Lewis.
\newblock Train short, test long: Attention with linear biases enables input
  length extrapolation, 2022.

\bibitem{sun2022lengthextrapolatable}
Yutao Sun, Li~Dong, Barun Patra, Shuming Ma, Shaohan Huang, Alon Benhaim,
  Vishrav Chaudhary, Xia Song, and Furu Wei.
\newblock A length-extrapolatable transformer, 2022.

\bibitem{openai2023gpt4}
OpenAI.
\newblock Gpt-4 technical report, 2023.

\bibitem{claudeanthropic}
Anthropic.
\newblock Introducing claude, 2023.

\bibitem{touvron2023llama}
Hugo Touvron, Thibaut Lavril, Gautier Izacard, Xavier Martinet, Marie-Anne
  Lachaux, Timothée Lacroix, Baptiste Rozière, Naman Goyal, Eric Hambro,
  Faisal Azhar, Aurelien Rodriguez, Armand Joulin, Edouard Grave, and Guillaume
  Lample.
\newblock Llama: Open and efficient foundation language models, 2023.

\bibitem{touvron2023llama2}
Hugo Touvron, Louis Martin, Kevin Stone, Peter Albert, Amjad Almahairi, Yasmine
  Babaei, Nikolay Bashlykov, Soumya Batra, Prajjwal Bhargava, Shruti Bhosale,
  Dan Bikel, Lukas Blecher, Cristian~Canton Ferrer, Moya Chen, Guillem
  Cucurull, David Esiobu, Jude Fernandes, Jeremy Fu, Wenyin Fu, Brian Fuller,
  Cynthia Gao, Vedanuj Goswami, Naman Goyal, Anthony Hartshorn, Saghar
  Hosseini, Rui Hou, Hakan Inan, Marcin Kardas, Viktor Kerkez, Madian Khabsa,
  Isabel Kloumann, Artem Korenev, Punit~Singh Koura, Marie-Anne Lachaux,
  Thibaut Lavril, Jenya Lee, Diana Liskovich, Yinghai Lu, Yuning Mao, Xavier
  Martinet, Todor Mihaylov, Pushkar Mishra, Igor Molybog, Yixin Nie, Andrew
  Poulton, Jeremy Reizenstein, Rashi Rungta, Kalyan Saladi, Alan Schelten, Ruan
  Silva, Eric~Michael Smith, Ranjan Subramanian, Xiaoqing~Ellen Tan, Binh Tang,
  Ross Taylor, Adina Williams, Jian~Xiang Kuan, Puxin Xu, Zheng Yan, Iliyan
  Zarov, Yuchen Zhang, Angela Fan, Melanie Kambadur, Sharan Narang, Aurelien
  Rodriguez, Robert Stojnic, Sergey Edunov, and Thomas Scialom.
\newblock Llama 2: Open foundation and fine-tuned chat models, 2023.

\bibitem{naturalquestions}
Tom Kwiatkowski, Jennimaria Palomaki, Olivia Redfield, Michael Collins, Ankur
  Parikh, Chris Alberti, Danielle Epstein, Illia Polosukhin, Matthew Kelcey,
  Jacob Devlin, Kenton Lee, Kristina~N. Toutanova, Llion Jones, Ming-Wei Chang,
  Andrew Dai, Jakob Uszkoreit, Quoc Le, and Slav Petrov.
\newblock Natural questions: a benchmark for question answering research.
\newblock {\em Transactions of the Association of Computational Linguistics},
  2019.

\bibitem{su2022roformer}
Jianlin Su, Yu~Lu, Shengfeng Pan, Ahmed Murtadha, Bo~Wen, and Yunfeng Liu.
\newblock Roformer: Enhanced transformer with rotary position embedding, 2022.

\bibitem{shaw2018selfattention}
Peter Shaw, Jakob Uszkoreit, and Ashish Vaswani.
\newblock Self-attention with relative position representations, 2018.

\bibitem{lmsysarena}
Lianmin Zheng, Wei-Lin Chiang, Ying Sheng, Siyuan Zhuang, Zhanghao Wu, Yonghao
  Zhuang, Zi~Lin, Zhuohan Li, Dacheng Li, Eric.~P Xing, Hao Zhang, Joseph~E.
  Gonzalez, and Ion Stoica.
\newblock Judging llm-as-a-judge with mt-bench and chatbot arena, 2023.

\bibitem{kaiokendev}
Things i’m learning while training superhot, 2023.

\bibitem{chen2023extending}
Shouyuan Chen, Sherman Wong, Liangjian Chen, and Yuandong Tian.
\newblock Extending context window of large language models via positional
  interpolation, 2023.

\bibitem{xu2021neural}
Keyulu Xu, Mozhi Zhang, Jingling Li, Simon~S. Du, Ken ichi Kawarabayashi, and
  Stefanie Jegelka.
\newblock How neural networks extrapolate: From feedforward to graph neural
  networks, 2021.

\bibitem{ruoss2023randomised}
Anian Ruoss, Grégoire Delétang, Tim Genewein, Jordi Grau-Moya, Róbert
  Csordás, Mehdi Bennani, Shane Legg, and Joel Veness.
\newblock Randomized positional encodings boost length generalization of
  transformers, 2023.

\bibitem{littleretrieval}
anadim.
\newblock A little retrieval test for large language models, May 2023.

\bibitem{longchat2023}
Dacheng Li*, Rulin Shao*, Anze Xie, Ying Sheng, Lianmin Zheng, Joseph~E.
  Gonzalez, Ion Stoica, Xuezhe Ma, and Hao Zhang.
\newblock How long can open-source llms truly promise on context length?, June
  2023.

\bibitem{liu2023lost}
Nelson~F. Liu, Kevin Lin, John Hewitt, Ashwin Paranjape, Michele Bevilacqua,
  Fabio Petroni, and Percy Liang.
\newblock Lost in the middle: How language models use long contexts, 2023.

\bibitem{zheng2023judging}
Lianmin Zheng, Wei-Lin Chiang, Ying Sheng, Siyuan Zhuang, Zhanghao Wu, Yonghao
  Zhuang, Zi~Lin, Zhuohan Li, Dacheng Li, Eric.~P Xing, Hao Zhang, Joseph~E.
  Gonzalez, and Ion Stoica.
\newblock Judging llm-as-a-judge with mt-bench and chatbot arena, 2023.

\bibitem{hu2021lora}
Edward~J. Hu, Yelong Shen, Phillip Wallis, Zeyuan Allen-Zhu, Yuanzhi Li, Shean
  Wang, Lu~Wang, and Weizhu Chen.
\newblock Lora: Low-rank adaptation of large language models, 2021.

\end{thebibliography}
\bibliographystyle{unsrt} 
\newpage
\appendix

\section{LLaMA 2}
\label{sec:llama-2}

As we were finalising this paper, Meta released LLaMA 2 \cite{touvron2023llama2}. We verified that similar results of context length extrapolation were achievable with LLaMA 2 by the linear interpolation method. We applied the same method as described in Section \ref{sec:experiment-setup}, training LLaMA 2-13B on a portion of the RedPajama dataset modified such that each data sample has a size of exactly 4096 tokens. We then also applied instruction finetuning with the Vicuna dataset. We used a scale of 8. Performance is shown in tables \ref{table:llama-2-fine-grained} and \ref{table:llama-2-qa}.

We see that the model is able to achieve non-zero accuracies on LongChat-Lines up to a context length of 22000, further than any of the models we tested in the main paper. The model is also able to achieve non-zero performance on the WikiQA variants up to 32k context. However, we do see diminishing accuracy in both tasks as the context length increases. It is also worth noting that the accuracies on both tasks are slightly lower than LLaMA 1 with scale 16 on the context lengths which both models are capable of producing non-zero results.

\begin{table}[!t]
\small
\centering
\begin{tabular}{c|c}
Context Length & LLaMA 2 Linear (8) \\
\hline
2500  & 0.48 \\
3600  & 0.42 \\
4200  & 0.32 \\
4800  & 0.74 \\
7100  & 0.56 \\
9400  & 0.50 \\
11800 & 0.50 \\
14000 & 0.42 \\
16000 & 0.38 \\
17500 & 0.14 \\
20000 & 0.14 \\
22000 & 0.08 \\
26000 & 0.00 \\
30000 & 0.00 \\
\end{tabular}
\caption{LLaMA 2 performance on LongChat-Lines with a scale factor of 8.}
\label{table:llama-2-fine-grained}
\end{table}

\begin{table}[!t]
\small
\centering
\begin{tabular}{c|cc}
Context Length & \multicolumn{2}{c}{LLaMA 2 Linear (8)} \\
               & AltQA & FFQA \\
\hline
\hline
2k   & 0.72 & 0.56 \\
4k   & 0.76 & 0.55 \\
8k   & 0.71 & 0.56 \\
16k  & 0.59 & 0.44 \\
24k  & 0.36 & 0.28 \\
32k  & 0.15 & 0.10 \\
\hline
\end{tabular}
\caption{LLaMA 2 performance on the WikiQA variants with a scale factor of 8.}
\label{table:llama-2-qa}
\end{table}

\newpage
\section{Loss Curves}
\label{sec:loss-curves}

\begin{figure}[h]
    \centering
    \includegraphics[width=0.5\textwidth]{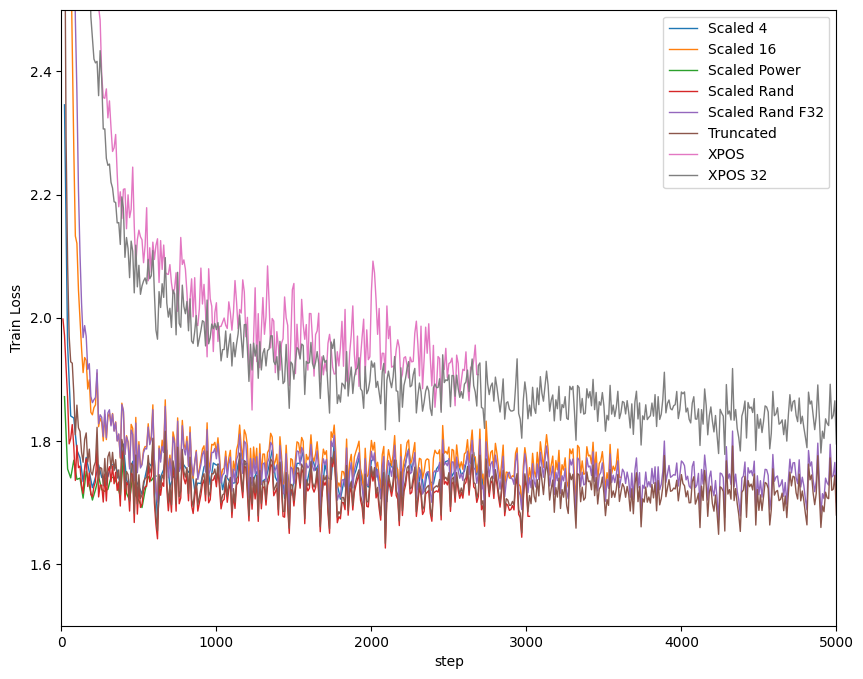}
    \caption{Training loss curves of models during the initial fitting runs on 4096 token samples extracted from the RedPajama dataset.}
    \label{fig:train-loss-curves}
\end{figure}

\begin{figure}[h]
    \centering
    \includegraphics[width=0.5\textwidth]{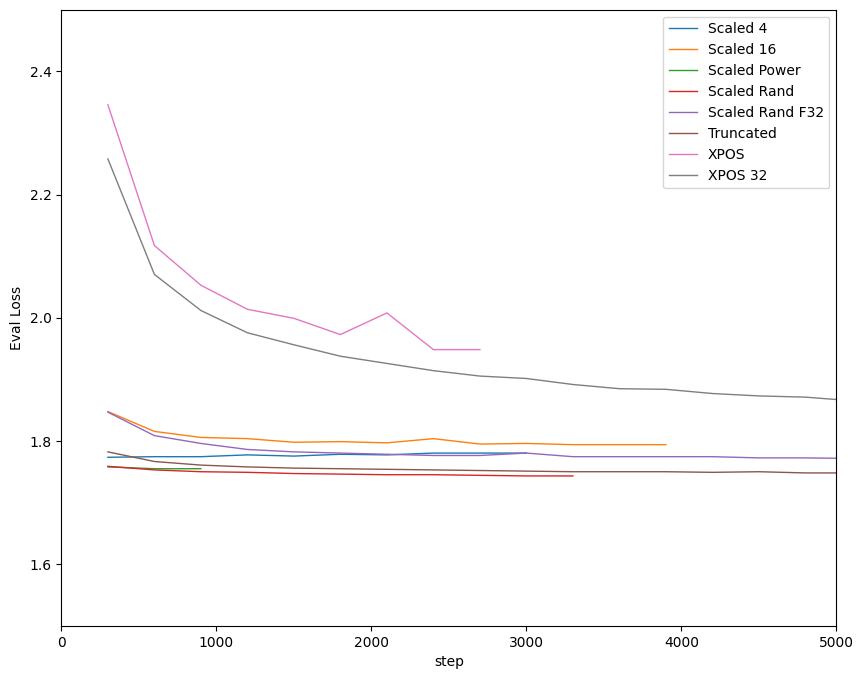}
    \caption{Evaluation loss curves of models during the initial fitting runs on 4096 token samples extracted from the RedPajama dataset.}
    \label{fig:eval-loss-curves}
\end{figure}

\newpage
\section{Impact of IFT}
\label{sec:impact-of-ift}

\begin{figure}[!t]
    \centering
    \includegraphics[width=0.6\textwidth]{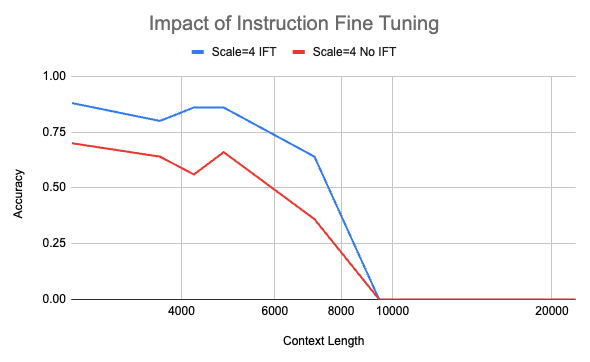}
    \caption{Comparison between IFT and non-IFT on LongChat-Lines with linear scaling of 4 applied. Although IFT improves the accuracies, it does not extend the range of contexts on which the model obtains a non-zero accuracy.}
    \label{fig:impact_of_ift}
\end{figure}

As mentioned in Section \ref{sec:results} of the main text, we found that instruction-fine-tuning with the Vicuna dataset did improve accuracies on LongChat-Lines, but did not change the span of non-zero contexts for a given model. Figure \ref{fig:impact_of_ift} shows this on the model with linear interpolation with scale 4.

\end{document}